\documentclass[sigconf]{acmart}
\usepackage{bm}
\usepackage{multirow}
\usepackage{amsmath}
\usepackage{gensymb}
\usepackage{textcomp}
\usepackage{chngcntr}
\usepackage{algorithm}
\usepackage{algpseudocode}
\usepackage{fancyhdr}
\usepackage{soul}
\usepackage[titletoc]{appendix}

\counterwithout{equation}{section} 

\AtBeginDocument{%
  }

\begin{document}
\title{Fine-gained air quality inference based on low-quality sensing data using self-supervised learning}

\author{Meng Xu}
\affiliation{%
	\institution{School of Transportation and Logistics at Southwest
		Jiaotong University}
	\city{Chengdu}
	\country{China}
}

\author{Ke Han}
\affiliation{%
	\institution{School of Economics and Management, Southwest Jiaotong University}
	\city{Chengdu}
	\country{China}
}
\email{kehan@swjtu.edu.cn}
\authornotemark[1]

\author{Weijian Hu}
\affiliation{%
	\institution{School of Transportation and Logistics at Southwest
		Jiaotong University}
	\city{Chengdu}
	\country{China}
}
\affiliation{%
	\institution{School of Digital and Intelligent Industry, Inner Mongolia University of Science and Technology}
	\city{Baotou}
	\country{China}
}

\author{Wen Ji}
\affiliation{%
	\institution{School of Transportation and Logistics at Southwest
		Jiaotong University}
	\city{Chengdu}
	\country{China}
}

\renewcommand\footnotetextcopyrightpermission[1]{}
\settopmatter{printacmref=false}
\renewcommand{\shortauthors}{Xu et al.}

\begin{abstract}
Fine-grained air quality (AQ) mapping is made possible by the proliferation of cheap AQ micro-stations (MSs). However, their measurements are often inaccurate and sensitive to local disturbances, in contrast to standardized stations (SSs) that provide accurate readings but fall short in number. To simultaneously address the issues of low data quality (MSs) and high label sparsity (SSs), a multi-task spatio-temporal network (MTSTN) is proposed, which employs self-supervised learning to utilize massive unlabeled data, aided by seasonal and trend decomposition of MS data offering reliable information as features. The MTSTN is applied to infer NO$_2$, O$_3$ and PM$_{2.5}$ concentrations in a 250 km$^2$ area in Chengdu, China, at a resolution of 500m$\times$500m$\times$1hr. Data from 55 SSs and 323 MSs were used, along with meteorological, traffic, geographic and timestamp data as features. The MTSTN excels in accuracy compared to several benchmarks, and its performance is greatly enhanced by utilizing low-quality MS data. A series of ablation and pressure tests demonstrate the results' robustness and interpretability, showcasing the MTSTN's practical value for accurate and affordable AQ inference. 
\end{abstract}


\maketitle
\pagestyle{fancy}
\fancyhf{}

\section{Introduction}
As the serious health hazards of air pollution are gradually revealed \cite{MXCRFAASMP2021, AD2014}, public demand for air quality improvement has become increasingly urgent. Fine-grained air quality information, crucial for air quality management, has received worldwide attention across academia and industry. However, obtaining this information through direct measurement is practically challenging. While standardized stations provide accurate and reliable measurements, their deployment is spatially sparse due to the high costs of construction and maintenance (\$100,000$\sim$300,000 for construction and over \$100,000 per year for maintenance) \cite{AMGB2017}. In contrast, cheap micro-stations can be ubiquitously deployed, offering a possible solution, but their readings are often inaccurate and sensitive to local disturbances. Consequently, fusing these two types of data for accurate and affordable air quality inference has significant scientific and practical values, which is addressed in this work.

Air quality inference is a challenging problem due to the complex and dynamic spatio-temporal dependencies of pollutant concentrations on various factors. Additionally, it is a few-labeled problem owing to the limited number of standardized stations. Over the past decade, spatio-temporal neural networks have gained popularity in this area for their remarkable competence in learning spatio-temporal dependencies. Among them, the most prevalent are the various kinds of supervised learning models \cite{HJYZLYR2023, CWYYL2018, HJYHYZR2023}. However, these models fail to address the sparsity of labels, underutilizing unlabeled data that constitute a majority portion. To overcome this issue, many researchers have pivoted towards semi-supervised learning models \cite{CLYYMCG2016, ZYFH2013}, which leverage both labeled and unlabeled data during the training process. Self-supervised learning models also demonstrate considerable competence in addressing few-labeled challenge \cite{RDAFMSFPMB2018, JWTHYSZJ2020}, but exhibit a gap in exploring air quality inference so far. Consequently, self-supervised learning models are novel and promising tools for air quality inference.


In this work, a self-supervised learning framework, called Multi-Task Spatio-Temporal Network (MTSTN), is proposed for fine-grained air quality inference over the graph structure. The MTSTN involves a self-supervised task and a supervised task; the former acts as a pretext task, learning a valuable representation from unlabeled data, while the latter serves as the downstream task, executing the core inference. The self-supervised learning is designed as a regression inference task, utilizing the spatial interpolation results based on high-quality data from standardized stations as its labels. On the use of low-quality data, we employ Seasonal and Trend Decomposition using Loess (STL) \cite{CRWJI1990} to extract relatively stable and reliable information from micro-station data, which is then used as features for self-supervised learning. Among the decomposition results, the trend exhibits high importance to the inference performance, showcasing the value of low-quality pervasive sensing data for accurate AQ inference. The main contributions of this paper are summarized as follows:

\begin{itemize}
	\item This work explores the hidden values of low-quality data provided by cheap and pervasive AQ sensors, simultaneously addressing the issues of low data quality (offered by micro-stations) and high label sparsity (offered by standardized stations), which has significant practical values for accurate and affordable air quality inference.  
	
	\item To achieve the above goal, we propose a novel Multi-Task Spatio-Temporal Network (MTSTN), which employs self-supervised learning to utilize massive unlabeled data, aided by seasonal and trend decomposition of micro-station data offering reliable spatio-temporal information as features. Compared to state-of-the-art baselines, this method results in higher inference accuracy.
	
	\item A method for the selection of both continuous and categorical features and importance assessment is proposed, by incorporating their gradients as regularization terms in the loss function, which further improves the model's accuracy and interpretability. Notably, in a real-world case study, trend from micro-station data decomposition turns out to be the most significant feature, showcasing the practical value of low-quality yet pervasive sensing data for accurate air quality inference. 
\end{itemize}


\section{Results}
\subsection{Problem Formalization}
Given the air quality graph $\mathcal{G}$, which represents the spatial position information of the study area, and two types of features: Grid features $\chi$ and graph topologies $\boldsymbol{A}$, as defined in \ref{Dataset and features}, the objective of air quality inference is to learn a function $\mathcal{F}$ that can accurately infer pollutant concentration $Y_{t}$. In summary, the air quality inference problem can be formulated as follows:

\begin{equation}
	Y_{t}=\mathcal{F} \left[ \chi_{\left( t-\tau +1 \right)},...,\chi_t;\boldsymbol{A} ; \mathcal{G} \right] 
\end{equation}
where $t$ and $\tau$ are timestamp and time window respectively.
\subsection{Dataset and features}
\begin{figure}[h]
	\centering
    \includegraphics[width=\linewidth]{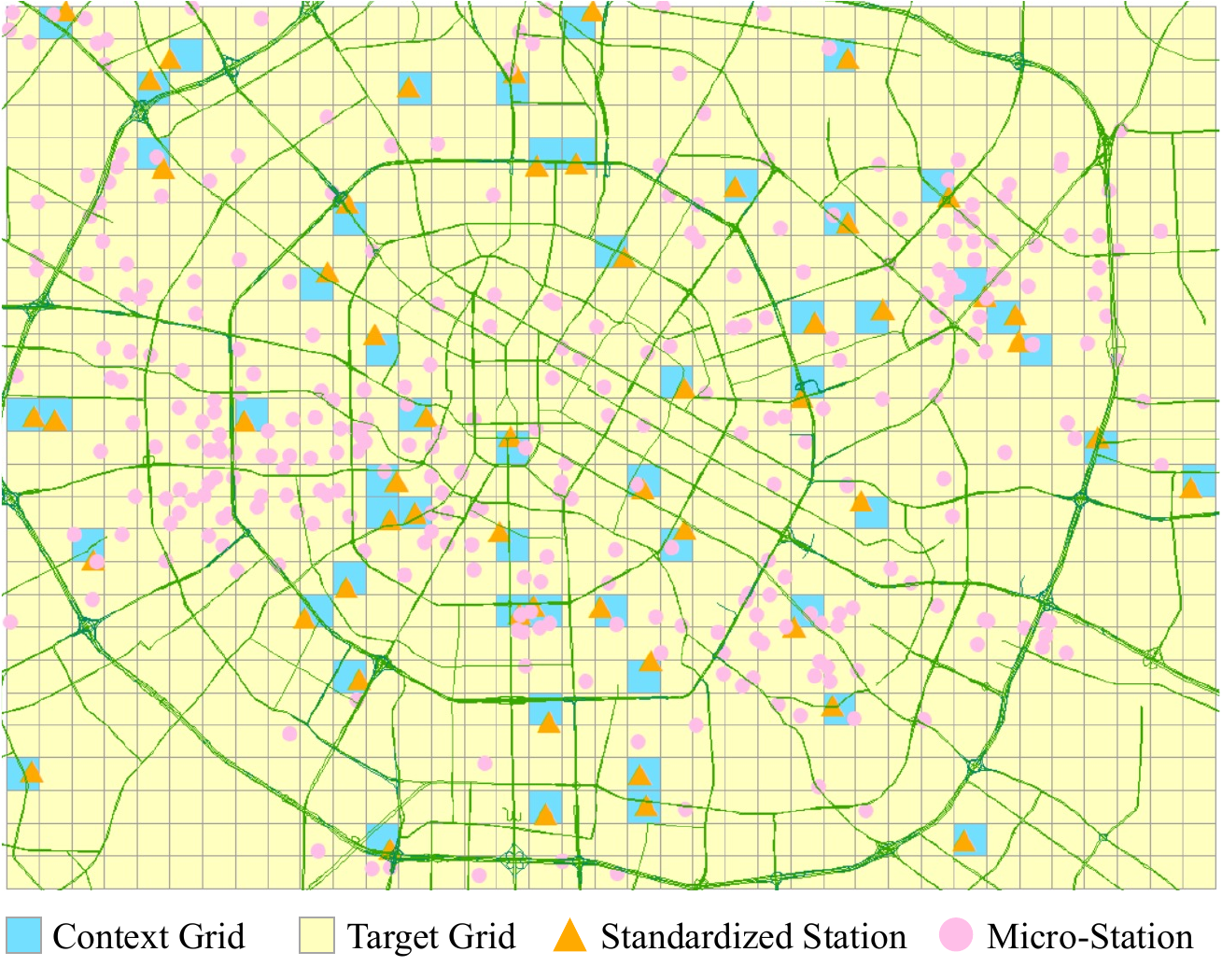}
	\caption{Distribution of 378 air quality monitoring stations (55 standardized stations and 323 micro-stations) across the central region of Chengdu, China. The context (labeled) grids have standardized stations, while the target (unlabeled) grids do not.}
	\label{map}
\end{figure}

This is based on a real-world dataset in the central region (250 km$^2$) of Chengdu, China, between 1st Mar. and 1st Apr. 2022. Space is meshed into 999 grids of size 500m$\times$500m (as depicted in Fig. \ref{map}), and time is discretized into 1-hr intervals. Additionally, 5 types of datasets are collected: hourly pollutant (NO$_2$, O$_3$, PM$_{2.5}$) concentrations from standardized and micro-stations, hourly meteorological data, traffic data (hourly road congestion index and truck GPS trajectories), geographic data from OpenStreetMap\footnote{\url{https://openmaptiles.org/languages/zh/}}, and timestamp data. Based on these datasets, 5 types of features are designed, with a total number of 42. Detailed descriptions and visualizations of other features are provided in \ref{Dataset and features}.

\begin{figure*}[h!]
	\centering
	\includegraphics[width=0.9\linewidth]{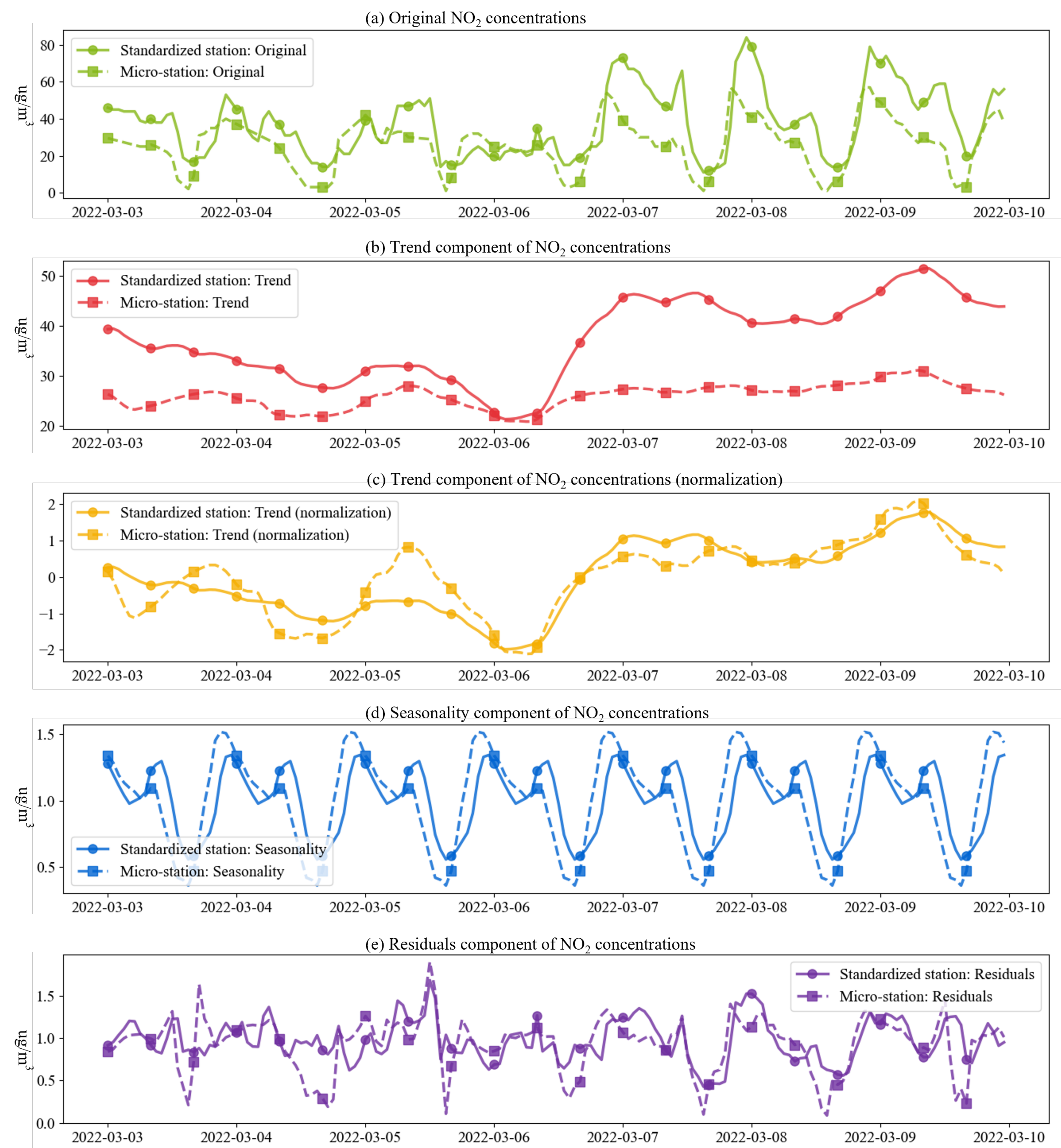}
	\caption{NO$_2$ concentrations and their decomposition results during Mar. 3 - 10, 2022}
	\label{STL}
\end{figure*}

As a hallmark of this work, the decomposition of micro-station data (NO$_2$ concentration) is illustrated. Fig. \ref{STL} (a) shows the NO$_2$ concentrations from both standardized and micro-stations within the same grid. It is apparent that the two NO$_2$ concentrations differ numerically, but exhibit a non-linear correlation. Therefore, before utilizing the pollutant concentrations collected by micro-stations, it is crucial to extract a valuable component from them. To achieve this, Seasonal and Trend Decomposition using Loess (STL) \cite{CRWJI1990} is applied to mine pollutant concentrations, decomposing them into trend, seasonality and residuals, as shown in Fig. \ref{STL} (b), (d) and (e). The decomposition results reveal similarity in trend and seasonality from the two data sources. Seasonality indicates that NO$_2$ concentrations increase nocturnally, likely due to the atmospheric titration effect, where lower temperatures and reduced solar radiation at night enhance the reaction between NO and O$_3$, leading to increased NO$_2$ formation. This finding supports the reliability of the decomposition results. To avoid feature redundancy, seasonality is not employed as a feature because other features (e.g. timestamps) can better indicate the periodicity of pollutant concentrations. Fig. \ref{STL} (b) shows highly correlated trends, but with distinct numerical values due to the sensors' characteristics. To overcome this issues, we normalize their trend, as demonstrated in Fig. \ref{STL} (c). After normalization, the numerical values of the two trends demonstrate a high degree of consistency. Consequently, normalized trend (abbreviated as trend) is employed as a feature for air quality inference.

\begin{figure}[h]
	\centering
	\includegraphics[width=1.0\linewidth]{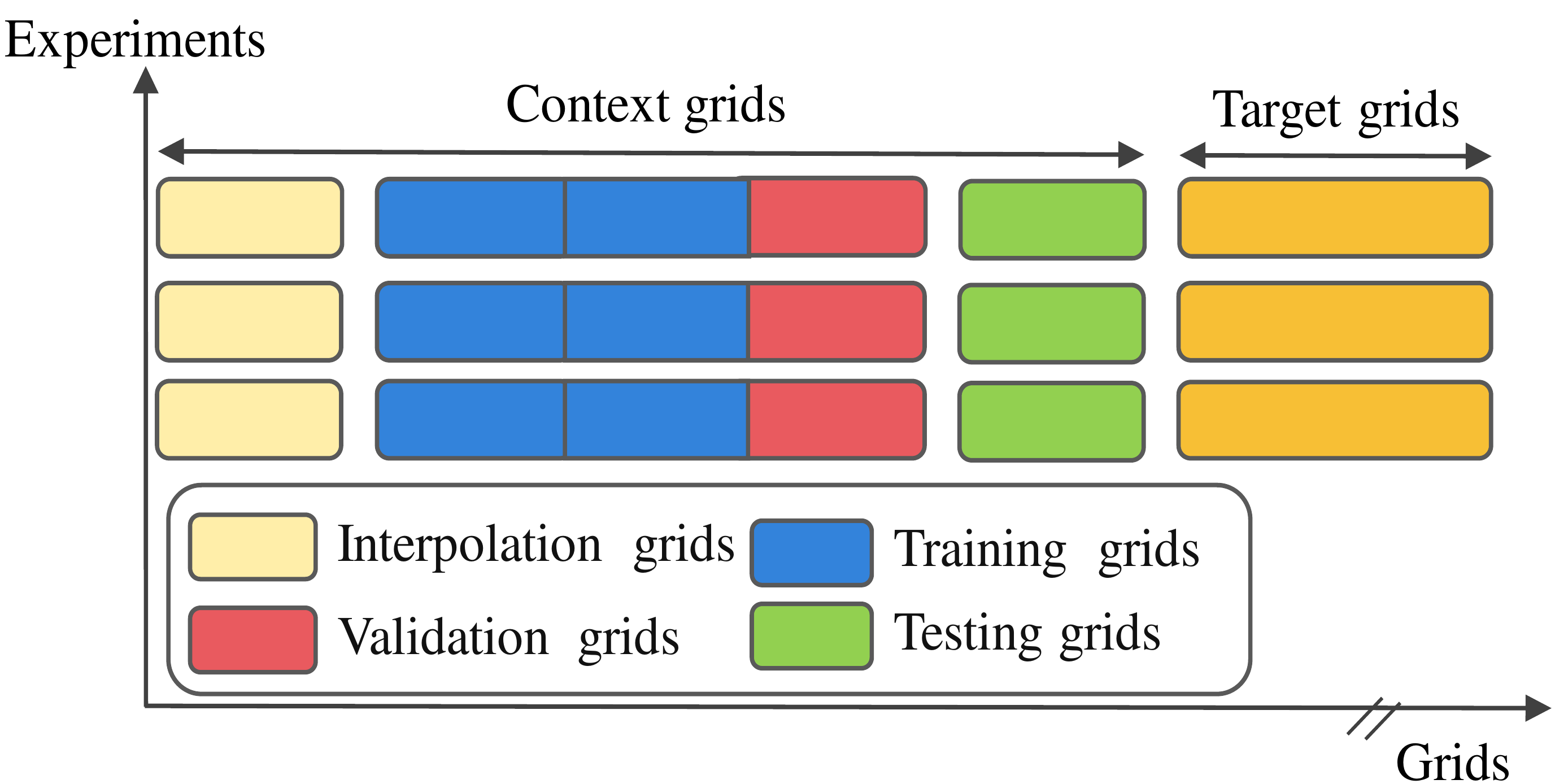}
	\caption{Evaluation strategy and dataset segmentation.}
	\label{evaluation}
\end{figure}

As demonstrated in Fig. \ref{evaluation}, the study area is segmented into four parts: interpolation grids, training grids, validation grids and test grids, which are designed to generate labels for self-supervised tasks, train the model, select hyperparameters and evaluate model performance respectively. Due to the few-labeled issue, only 20\% of the context grids are set as interpolation grids and 10\% are set as test grids. For the remaining 70\% of the context grids, a 3-fold cross-validation is implemented to reduce parameter selection randomness and improve model confidence. The average evaluation metrics on the test grids are employed to quantify model performance, including mean absolute error (MAE), root mean square errror (RMSE) and r-square (R$^2$), which are commonly used in regression problems. To validate MTSTN's generalizability in pollutant concentration inference, experiments are conducted on three pollutants: NO$_2$, O$_3$ and PM$_{2.5}$. 

\subsection{Loss function}
To implement both continuous and categorical feature selection, two gradient-based regularization terms are incorporated into the loss function, formulated as follows:
\begin{equation}
	\begin{aligned}
		\theta ^*,\varTheta ^*,\varTheta _{\text{ss}}^{*}&=\underset{\theta ,\varTheta ,\varTheta _{\text{ss}}}{\text{argmin}}\left( 1-\beta -\gamma \right) \left( \alpha _{\text{sup}}\mathcal{L} _{\left( \theta ,\varTheta \right)}+\alpha _{\text{ss}}\mathcal{L} _{\left( \theta ,\varTheta _{\text{ss}} \right)} \right) \\
		&+\beta \sum_{i=1}^U{\left\| \frac{\partial \mathcal{L} \left( X _{\text{num}},\hat{Y} \right)}{\partial X _{\text{num}}^{i}} \right\| _2+}\gamma \sum_{j=1}^V{\sum_{q=1}^{Q_j}{\left\| \frac{\partial \mathcal{L} \left( X _{\text{cat}},\hat{Y} \right)}{\partial X _{\text{cat}}^{q}} \right\| _2}} \\
		\mathcal{L} &=\frac{1}{\left| Y_{\text{train}} \right|}\sum_{y_n\in Y_{\text{train}}}{\left| y_n-\hat{y}_{n} \right|}
	\end{aligned}
\end{equation}
where $\theta ^*,\varTheta ^*,\varTheta _{\text{ss}}^{*}$ are learnable parameters. $\alpha _{\text{sup}},\alpha _{\text{ss}}\in \mathbb{R} _{>0}$ serve as the loss weights for the supervised and self-supervised task respectively, which affect the role of each task on model's training process. $\beta $ and $\gamma $ are regularization coefficients for feature selection.
$U$ and $V$ are the numbers of continuous and categorical features respectively. $Q_j$ is the embedding dimension corresponding to the $j^{th}$ categorical feature. $\mathcal{L}$ is the function quantifying the discrepancy between ground truth $y_n$ and inference value $\hat{y}_{n}$ within the training dataset $Y{\text{train}}$.

\subsection{Model configurations}
All experiments are implemented with PyTorch 2.1 and trained on an NVIDIA GeForce RTX 4090 GPU. The loss function is minimized by Adam optimizer. To prevent overfitting and optimize training efficiency, the training process is allowed to be early stopped according to the performance on validation grids. More details about experiment settings and hyperparameters can be found in \ref{Setting}.

\subsection{Baselines}
Several baselines with different architectures are selected to compare their performances with MTSTN, including K-Nearest Neighbors (KNN), Inverse Distance Weighting (IDW), Land Use Regression (LUR), Random Forest (RF), Extreme Gradient Boosting (XGBoost), Spatio-Temporal Graph Convolutional Networks (STGCN), Multi-View Spatial-Temporal Graph Convolutional Networks (MSTGCN), Attention-based Spatial-Temporal Graph Neural Networks (ASTGNN) and Propagation Delay-Aware Dynamic Long-Range Transformer (PDFormer). More details about baselines can be found in \ref{Hyperparameter study}. For a fair comparison, different hyperparameters are tuned for each baseline to find their optimal settings.

\subsection{Overall performance}
Table \ref{performance} reports the overall performance of MTSTN and other baselines on the dataset. The MTSTN significantly outperforms all baselines across nearly every metric, demonstrating its superiority. Moreover, among the three pollutants, PM$_{2.5}$ has the lowest inference accuracy, primarily attributed to its high variability. This variability is mainly caused by the complexity of its sources and composition, as well as its extremely fine particle size, which complicates monitoring efforts. In addition, compared to certain deep learning models (STGCN, MSTGCN, ASTGNN, and PDFormer), traditional decision tree models such as RF and XGBoost demonstrate superior performance, highlighting that deep learning models do not consistently outperform decision tree models \cite{GYIVA2021}. 

\begin{table*}[h]
	\caption{Performance comparison between MTSTN and the baselines. The bold/underlined fonts mean the best/the second best results. Benchmarked performance denotes the degree of change in MTSTN relative to the best baseline, where $\uparrow$ indicates performance improvement, and $\downarrow$ signifies performance deterioration.}
	\label{performance}
	\begin{tabular}{cccccccccc}
		\toprule
		\multirow{2}{*}{Model}
		&\multicolumn{3}{c}{NO$_2$} & \multicolumn{3}{c}{O$_3$} & \multicolumn{3}{c}{PM$_{2.5}$}\\
		\cmidrule(r){2-4} \cmidrule(r){5-7} \cmidrule{8-10}
		& MAE&RMSE&R$^2$&MAE&RMSE&R$^2$&MAE&RMSE&R$^2$\\
		\midrule
		KNN (Hu \textit{et al.} \cite{CT1968}, 1968) &6.806&9.400&0.668&4.303&6.917&0.856&11.137&15.082&0.755\\
		IDW (Bartier \textit{et al.} \cite{BPC1996}, 1996)&6.741&9.401&0.668&4.273&7.347&0.838&11.357&15.912&0.725\\
		LUR (Briggs \textit{et al.} \cite{BDSPPSEKHK1997}, 1997)&7.193&9.906&0.632&6.526&10.105&0.696&12.369&15.555&0.739\\ 
		RF (Breiman \textit{et al.} \cite{BL2001}, 2001)&$\underline{4.206}$&$\underline{5.933}$&$\underline{0.868}$&\textbf{2.881}&$\underline{4.691}$&$\underline{0.934}$&8.153&11.051&0.868\\
		XGBoost (Chen \textit{et al.} \cite{CTTMVYHKRIT2015}, 2015)&4.428&6.261&0.853&3.160&5.042&0.924&$\underline{8.070}$&\textbf{10.663}&$\underline{0.877}$\\
		STGCN (Yu \textit{et al.} \cite{YBHZ2017}, 2017)&5.594&7.466&0.796&3.901&5.993&0.891&8.378&11.376&0.855\\
		MSTGCN (Jia \textit{et al.} \cite{JZYJXYRYH2021}, 2021)&4.857&6.462&0.823&3.957&6.097&0.904&10.489&13.744&0.807\\
		ASTGNN (Guo \textit{et al.} \cite{GSYHX2021}, 2021)&5.860&7.634&0.789&5.594&9.644&0.723&15.191&11.344&0.763\\
		PDFormer (Jiang \textit{et al.} \cite{JJCWJ2023}, 2023)&5.326&7.289&0.809&5.231&7.755&0.839&10.281&13.565&0.810\\
		MTSTN (Proposed method, 2024)&\textbf{3.758} &\textbf{5.242}&\textbf{0.910} &$\underline{3.042}$&\textbf{4.574}&\textbf{0.947}&\textbf{7.767}&$\underline{10.666}$&\textbf{0.883}\\
		\midrule
		Benchmarked performance& $\uparrow$ 10.65\%& $\uparrow$ 11.65\%&$\uparrow$ 4.84\%&$\downarrow$ 5.59\%&$\uparrow$ 2.49\%&$\uparrow$ 1.39\%&$\uparrow$ 3.75\%&$\downarrow$ 0.03\%&$\uparrow$ 0.68\%\\
		\bottomrule
	\end{tabular}
\end{table*}

\subsection{Feature importance}
Feature selection and importance assessment are crucial for improving the model accuracy and interpretability. Feature importance is calculated based on the gradients, which guide the weights update in neural networks. A larger gradient value indicates greater importance of the feature. Since categorical features are embedded before being input into the feature encoder module, distinct methods are implemented to determine the importance of continuous and categorical features, which are formulated as follows:

\begin{equation}
	\begin{aligned}
		p_{\text{num}}^{i}&=\left\| \frac{\partial \mathcal{L} \left( X_{\text{num}},Y \right)}{\partial X_{\text{num}}^{i}} \right\| \\
		p_{\text{cat}}^{j}&=\frac{\sum_{q=1}^{Q_j}{\left\| \frac{\partial \mathcal{L} \left( X_{\text{cat}},Y \right)}{\partial X_{\text{cat}}^{q}} \right\|}}{Q_j}
	\end{aligned}
\end{equation}
where $p_{\text{num}}^{i}$ and $p_{\text{cat}}^{j}$ denote the importance of the $i^{\text{th}}$ numerical feature and the $j^{\text{th}}$ categorical feature respectively. 

Fig. \ref{key_feature} reports the rankings of top 20 features. The following are observed: a) The feature \emph{trend} takes the top position for all three pollutants, highlighting the significance of using low-quality data from micro-stations. b) For a given pollutant, other pollutants' concentrations are  significant features. For example, NO$_2$ and O$_3$ concentrations are important features to each other, confirming their chemical interaction (the titration effect); in addition, NO$_2$ concentration ranks 2$^{\text{nd}}$ for PM$_{2.5}$ inference, as the former may contributes to the nocturnal formation of nitrate particles, which are a significant component of PM$_{2.5}$ \cite{FZLHZXDCSF2021, YTNXWGMZHL2023}. c) Meteorological features take up over half of the top 10 rankings, suggesting their critical roles in pollutant formation and dissipation. d) Periodicity in the pollutant concentrations is represented by the feature \emph{hour of day}, which is ranked 10 (NO$_2$), 11 (O$_3$), and 16 (PM$_{2.5}$). Indeed, this is confirmed by Fig. S4. in \ref{Timestamp data}. e) \emph{Grid congestion index}, ranking 1$^{\text{st}}$ among all features related to human activities, implies that traffic congestion levels have a significant impact on these pollutant concentrations, highlighting the critical role of congestion mitigation in combating air pollution. 

\subsection{Ablation study}
To investigated the effectiveness of each core module in MTSTN, a set of ablation experiments are conducted. The experimental results are reported in Table \ref{Ablation experiments}

(1) \textbf{Effects of self-supervised task}: To evaluate the effectiveness of proposed self-supervised task, it is compared with three variants: a) w/o SST: removing self-supervised task; b) r/p KSST: replacing IDW with KNN. c) r/p GSST: replacing the self-supervised task with the graph completion task. Removing the self-supervised task leads to a degradation in MAE, RMSE and R$^2$, revealing the powerful learning capability of the self-supervised task. Additionally, r/p GSST exhibits inferior performance compared to MTSTN and w/o KSST, because the self-supervised task of r/p GSST is graph completion, while those of the latter two are spatio-temporal regression inference tasks which are analogous to the supervised task. 

(2) \textbf{Effects of positional embedding module}: The positional embedding module is removed (w/o PE) to examine its efficacy. Results indicate that the MTSTN's performance declines after the removal, highlighting the significance of grids' relative positional correlations for air quality inference.

\begin{figure*}[h]
	\centering
	\includegraphics[width=1.0\linewidth]{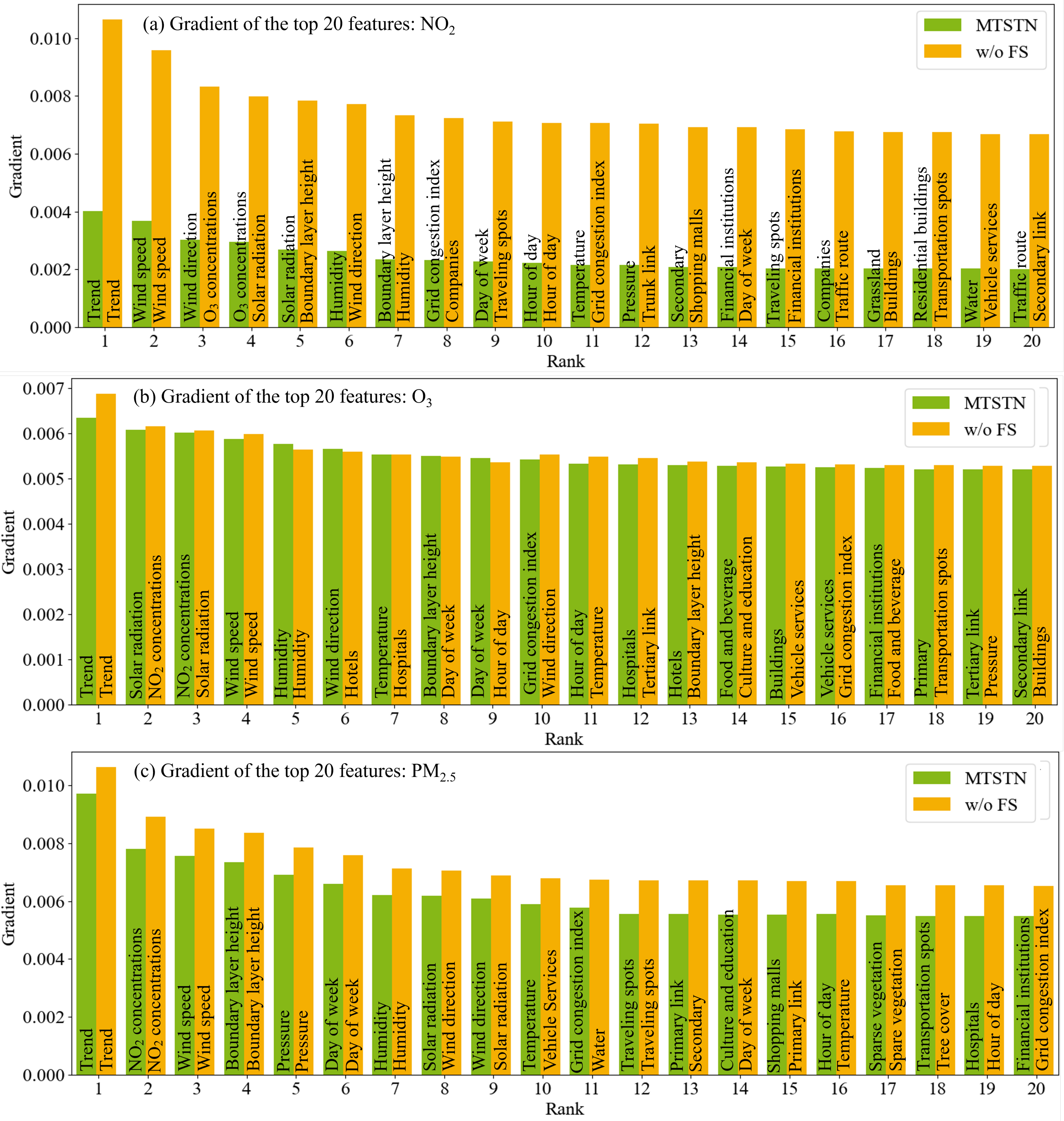}
	\caption{Key feature ranking}
	\label{key_feature}
\end{figure*}

(3) \textbf{Effects of feature selection}: A model is constructed without the proposed feature selection method  (w/o FS), whose deteriorated performance underscores the crucial role of feature selection in air quality inference. Fig. \ref{key_feature} illustrates the key features of both w/o FS and MTSTN. Using NO$_2$ as an example, temperature and pressure, both strongly correlated with NO$_2$ concentrations \cite{VACMMAB2002, GDSGAD2021}, are identified as key features in MTSTN but not in w/o FS. This shows better adherence to the natural laws of air pollution by the proposed feature selection method.

(4) \textbf{Effects of adjacency matrix}: In this work, two adjacency matrices: OD adjacency matrix ($A^{\text{OD}}$) and semantic adjacency matrix ($A^{\text{SE}}$), are employed as features. Further details about these matrices are available in \ref{Traffic data} and \ref{Geographic}. Two removal experiments are conducted for each matrix, named w/o OD and w/o SE. Experimental results demonstrate that the removal of either $A^{\text{OD}}$ or $A^{\text{SE}}$ degrades the performance across all evaluation metrics, underscoring their essential role in air quality inference. Compared to $A^{\text{OD}}$, the removal of $A^{\text{SE}}$ leads to a more pronounced performance degradation, suggesting that $A^{\text{SE}}$ has a more substantial impact on air quality inference.

\begin{table*}[h]
	\caption{Experimental results of ablation studies. The bold/underlined font mean the best/the second best result.}
	\label{Ablation experiments}
	\begin{tabular}{cccccccccc}
		\toprule
		\multirow{2}{*}{Model}
		&\multicolumn{3}{c}{NO$_2$} & \multicolumn{3}{c}{O$_3$} & \multicolumn{3}{c}{PM$_{2.5}$}\\
		\cmidrule(r){2-4} \cmidrule(r){5-7} \cmidrule{8-10}
		& MAE&RMSE&R$^2$&MAE&RMSE&R$^2$&MAE&RMSE&R$^2$\\
		\midrule
		MTSTN&\textbf{3.758}&\textbf{5.242}&\textbf{0.910}&\textbf{3.042}&\textbf{4.574}&\textbf{0.947}&\textbf{7.767}&\textbf{10.666}&\textbf{0.883}\\
		w/o SST&4.221&5.818&0.886&3.254&5.032&0.936&11.008&14.611&0.774\\
	    r/p KSST&4.031&5.701&$\underline{0.892}$&3.554&5.735&0.916&10.221&13.353&0.813\\ 
		r/p GSST&4.233&5.760&0.890&3.531&5.766&0.915&10.618&14.045&0.797\\
		w/o PE&$\underline{3.956}$&$\underline{5.501}$&0.886&3.405&6.355&0.862&8.259&11.198&0.860\\
		w/o FS&4.362&5.994&0.881&3.173&$\underline{4.842}$&$\underline{0.941}$&$\underline{8.004}$&$\underline{10.975}$&$\underline{0.876}$\\
		w/o OD&4.212&5.859&0.884&$\underline{3.140}$&5.121&0.925&8.194&11.510&0.855\\
		w/o SE&4.305&5.941&0.861&3.158&4.974&0.919&8.915&12.048&0.841\\
		\bottomrule
	\end{tabular}
\end{table*}

\subsection{Missing ratio study}
\begin{figure*}[h]
	\centering
	\includegraphics[width=1.0\linewidth]{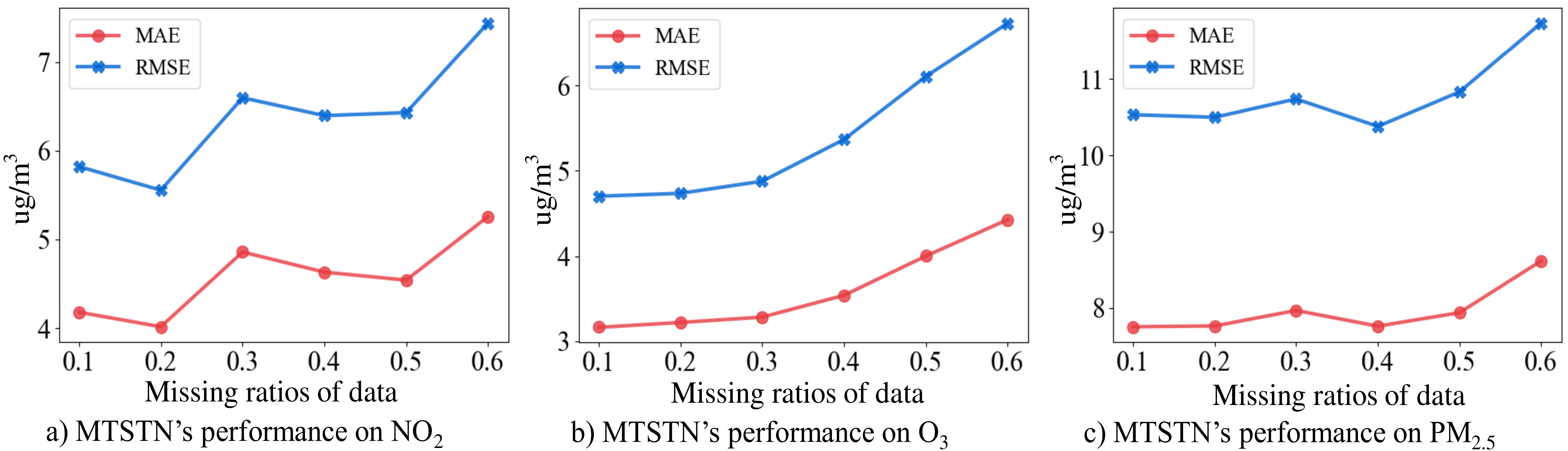}
	\caption{Data missing ratio study.}
	\label{missing ratio}
\end{figure*}

In reality, air quality monitoring stations often encounter signal loss due to maintenance or other unpredictable factors, making it essential to evaluate MTSTN's performance with missing data. Consequently, the data is randomly corrupted according to a predetermined missing ratio, and then restored using linear interpolation. For data points still missing after interpolation, values from the next timestamp within the same grid are used to fill them. Fig. \ref{missing ratio} reports the MAE and RMSE of MTSTN across missing ratios from 0.2 to 0.7, revealing that MTSTN's performances decline for all three pollutants due to errors caused by filling missing values, with a generally progressive decrease as the missing ratio increased. However, the degree of this decrease is minimal, demonstrating MTSTN's robustness. Notably, in real-world scenarios, the missing data ratio is typically low, so performance degradation remains within an acceptable range.

\subsection{A Case Study on NO$_2$}
\begin{figure*}[h]
	\centering
	\includegraphics[width=0.9\linewidth]{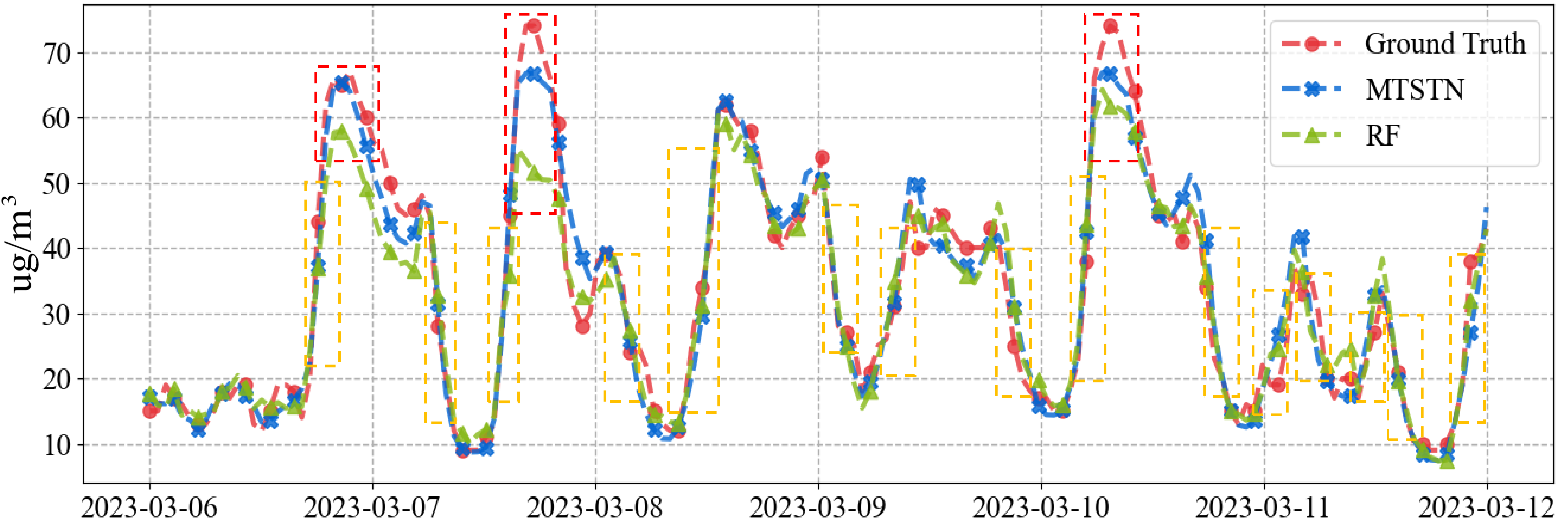}
	\caption{NO$_2$ concentration inference results on Chengdu's dataset during the Mar. 6 - 12, 2022}
	\label{temporal_visualisation}
\end{figure*}

\textbf{Temporal analysis}:
Fig. \ref{temporal_visualisation} provides evidence that MTSTN achieves superior inference accuracy compared to the second best baseline (RF) for the majority of the observation period. In particular, during the periods highlighted in the red box, i.e., the time when the NO$_2$ concentrations around the peak, inference results of MTSTN are consistently closer to the ground truth than those of RF. This superior performance can be attributed to that MTSTN uses spatial interpolation results as labels for the self-supervised task, which provide a reference value for the NO$_2$ concentrations. Additionally, both MTSTN and RF exhibit trends that closely align with the ground truth, especially when the trend is notable and steady, as indicated in the yellow box. This alignment may be due to the incorporation of trend data.

\begin{figure*}[h]
	\centering
	\includegraphics[width=1.0\linewidth]{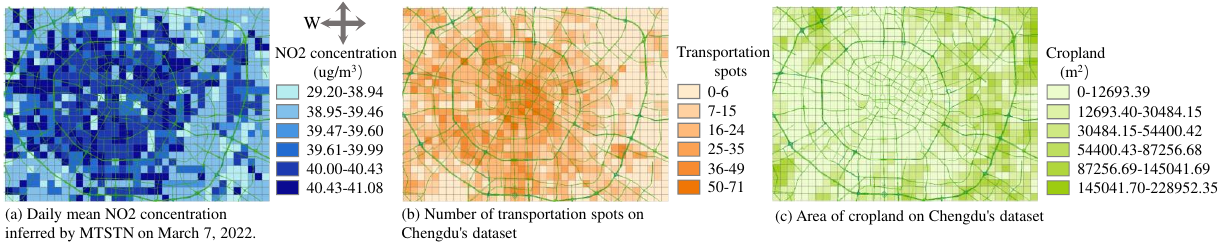}
	\caption{Spatial distribution of NO$_2$ concentration inference results, transportation spots and cropland. }
	\label{spatial_visualisation}
\end{figure*}

\textbf{Spatial analysis}: Fig. \ref{spatial_visualisation} (a) illustrates the spatial distribution of daily mean NO$_2$ concentrations, highlighting that the central region and the areas west of the center consistently exhibit higher NO$_2$ concentrations. Fig. \ref{spatial_visualisation} (b) and Fig. \ref{temporal_visualisation} (c) demonstrate the spatial distribution of traffic spots and cropland respectively. Traffic spots presents a similar spatial distribution to NO$_2$ concentrations, suggesting that traffic is a significant contributor to NO$_2$ emissions. While cropland shows a spatial distribution diametrically opposed to NO$_2$ concentrations. This contrast may be partly due to the soil in croplands effectively adsorbing NO$_2$ and partly because croplands are typically located in suburban areas where NO$_2$ emissions are generally lower.

\section{Discussion}
This paper proposes the Multi-Task Spatio-Temporal Network (MTSTN) for real-time and fine-grained air quality inference, which decouples the task into a supervised task and a self-supervised task. Notably, the self-supervised task can learn a valuable representation from unlabeled data, thus avoiding the data waste. For the readings collected by micro-stations, which are inaccurate and sensitive to local disturbances, Seasonal and Trend Decomposition using Loess (STL) is employed to extract their trend component, which is then utilized as a feature for inference. Furthermore, a gradient-based method for both continuous and categorical feature selection and importance assessment is proposed, effectively addressing the challenge of categorical feature selection and importance assessment. Experimental results demonstrate that MTSTN achieves significant performance improvement compared to other competitive baselines on Chengdu's dataset. The feature importance ranking demonstrates that trend is the most important feature, highlighting the practical value of low-data-quality pervasive sensors for high-accuracy air quality inference. 

\section{Methods}

\subsection{Learning framework}
\begin{figure*}[h]
	\centering
	\includegraphics[width=0.7\linewidth]{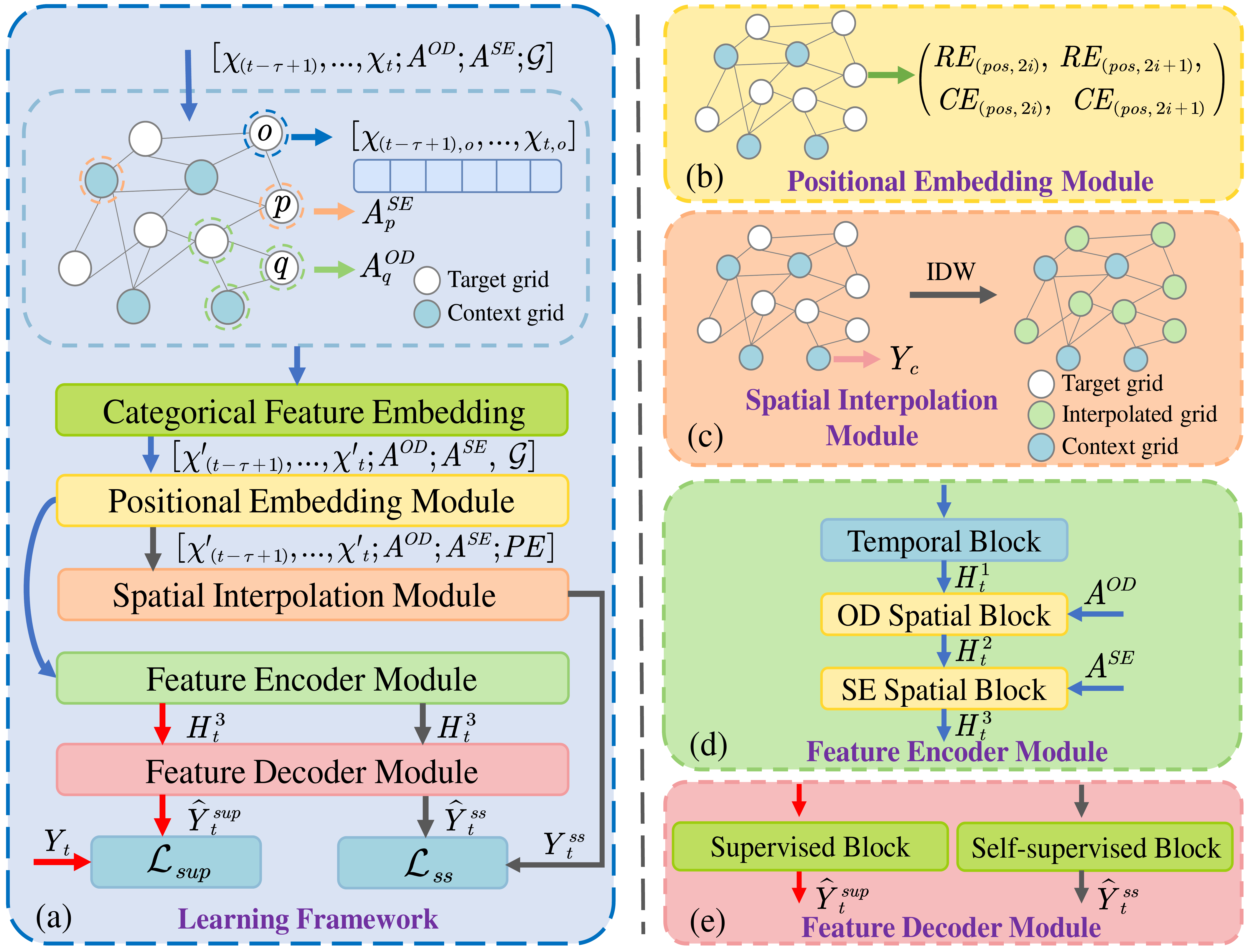}	
	\caption{The overall framework. Red arrows show the information flow pertinent to the supervised task, gray arrows indicate the information flow for the self-supervised task and blue arrows represent the shared components between both of them. (a) Learning framework of MTSTN. (b) Positional embedding module. (c) Spatial interpolation module. (d) Feature encoder module. (e) Feature decoder module.
	}
	\label{framework}
\end{figure*}

Inspired by You \textit{et al.} \cite{YCTZY2020}, a self-supervised learning framework called MTSTN is proposed, with the overall framework shown in Fig. \ref{framework}. MTSTN comprises four key modules: positional embedding module, spatial interpolation module, feature encoder module, and feature decoder module. Inference begins by processing the categorical features within the input $[ \chi_{\left( t-\tau +1 \right)},...,\chi_t;A^{\text{OD}};A^{\text{SE}};\mathcal{G} ]$ using categorical feature embedding. To capture relative position correlations, the embedding $
[ \chi ^{\prime}_{\left( t-\tau +1 \right)},...,\chi ^{\prime}_t;A^{\text{OD}};A^{\text{SE}};\mathcal{G} ] 
$ is processed through positional embedding module, producing a new embedding $[ \chi^{\prime}_{\left( t-\tau +1 \right)},...,\chi ^{\prime}_t;A^{\text{OD}};A^{\text{SE}};PE]$. This embedding is then used in two parallel pathways: one utilized by spatial interpolation module to generate labels $\hat{Y}_{t}^{ss}$ for the self-supervised learning, and the other employed by feature encoder module to learn a representation $H_{t}^{3}$. This representation is subsequently processed by feature decoder module, computing inference results for both self-supervised task and supervised task. 

\subsection{Positional embedding module}
Considering the robust correlation between pollutant concentrations and spatial position, a positional embedding module is incorporated to enhance the spatial awareness of MTSTN, as shown in Fig. \ref{framework} (b). Inspired by masked autoencoder \cite{HKXSY2022} and transformer \cite{VANNJLA2017}, sine and cosine functions are employed to construct positional embedding module. Specifically, positional embedding module consists of two distinct embeddings: row embedding (RE) and column embedding (CE), which can be represented as follows:

\begin{equation}
	\begin{aligned}
		RE_{\left( \text{pos}_r,2i \right)}&=\sin \left( pos_r/10000^{4i/d_p} \right) \\
		RE_{\left( \text{pos}_r,2i+1 \right)}&=\cos \left( pos_r/10000^{4i/d_p} \right)\\
		CE_{\left( \text{pos}_c,2i \right)}&=\sin \left( pos_c/10000^{4i/d_p} \right)\\
		CE_{\left( \text{pos}_c,2i+1 \right)}&=\cos \left( pos_c/10000^{4i/d_p} \right)
	\end{aligned}
\end{equation}
where $\text{pos}_r$ and $\text{pos}_c$ denote the row position and column position respectively, and $d_p$ signifies the dimension.

\subsection{Spatial interpolation module}
To learn a valuable representation, the self-supervised task should be thoughtfully designed \cite{SXWHQ2021}. Considering a self-supervised task analogous to the supervised task may improve the accuracy of air quality inference, this work structures the self-supervised task as a regression inference task, mirroring to the supervised task. As shown in Fig. \ref{framework} (c), Inverse Distance Weighting (IDW) (Eqn. \ref{IDW}) is used to spatially interpolate the pollutant concentrations, and the interpolation results then serve as the labels for the self-supervised task. IDW is renowned as a deterministic model for spatial interpolation, praised for its computational efficiency, ease of calculation, and clear interpretability \cite{LGD2008}. 

\begin{equation}\label{IDW}
	\begin{aligned}
		X_{g} =& \frac{1/d_{gc}^{p}}{\sum\nolimits_1^C{1/d_{gc}^{p}}} \times X_c\\
		d_{gc} = &2r \arcsin\sqrt{\sin^2\left(\frac{\phi_g - \phi_c}{2}\right) + \cos(\phi_c) \cos(\phi_g) \sin^2\left(\frac{\lambda_g - \lambda_c}{2}\right)}
	\end{aligned}
\end{equation}
where $X_{g}$ denotes the interpolation results for the target grid $r_g$. $d_{gc}^{k}$ represents the great-circle distance between the target grid $r_g$ and the context grid $r_c$, with $p$ as a constant. $r$ is the radius of the earth, i.e., $6371$ km, and both the longitude ($\phi_n$) and latitude ($\lambda_n$) of $r_n$ are expressed in radians. 

\subsection{Feature encoder module}
To capture complex and dynamic spatio-temporal dependencies of pollutant concentrations, the feature encoder module is designed as a spatio-temporal (ST) encoder module. As depicted in Fig. \ref{framework}(d), this module is structured with a temporal block followed by two spatial blocks. Notably, within the feature encoder module, the supervised task and the self-supervised task share parameters, facilitating information sharing across tasks and consequently improving the accuracy of air quality inference.
  
\emph{\textbf{Temporal Block}}. To capture the sequential patterns of pollutant concentrations across different timestamps, temporal block (TB) employs the bidirectional long-short term memory (Bi-LSTM) with the gated mechanism \cite{SMPK1997}, which preserves both past and future information. Specifically, TB takes $( \chi ^{\prime}_{( t-\tau +1 )},...,\chi ^{\prime}_t;PE)$ as input, and outputs a time-aware representation:

\begin{equation}
	H_{t}^{1}=TB\left( \chi ^{\prime}_{\left( t-\tau +1 \right)},...,\chi ^{\prime}_{t}; PE\right) 
\end{equation}
where $H_{t}^{1}=\{ \vec{h}_{t,1}^{1},...,\vec{h}_{t,N}^{1} \} \in \mathbb{R} ^{N\times d_t}$ denotes the time-aware representation at timestamp $t$, and $d_t$ represents the embedding dimension. 

Each LSTM, whether operating in the forward or backward direction, is equipped with three gates: forget (Eqn.\ref{forget}), input (Eqn.\ref{input}), and output (Eqn.\ref{output}) gates. These gates regulate the information flow, allowing LSTM to selectively preserve, refresh or eliminate information, thereby optimizing the feature learning process for complex temporal patterns.
\begin{equation}
	f_t=\sigma \left( W_f\left[ h_{t-1},\boldsymbol{\chi }\prime_{\boldsymbol{t}} \right] +b_f \right) 
	\label{forget}
\end{equation}
where $f_t$ is the forget gate, responsible for deciding whether to retain or discard the existing memory. $h_{t-1}$ denotes the hidden vector at timestamp $t-1$ and $\boldsymbol{\chi}_t^{\prime}=( \chi ^{\prime}_{t}; PE) $ signifies the input at timestamp $t$. $\sigma$ is the sigmoid function, while $W_{f}$ and $b_f$ are all learnable parameters.

\begin{equation}
	\begin{aligned}
	i_t&=\sigma \left( W_i\left[ h_{t-1},\boldsymbol{\chi }_{t}^{\prime} \right] +b_i \right)\\
	\tilde{C}_t&=\tanh \left( W_c\left[ h_{t-1},\boldsymbol{\chi }_{t}^{\prime} \right] +b_c \right)\\
	C_t&=f_tC_{t-1}+i_t\tilde{C}_t
\end{aligned}
	\label{input}
\end{equation}
where $i_t$ is the input gate, tasked with deciding whether to introduce new memory and what type of memory to introduce. $\tilde{C}_t$ indicates candidate memory, $C_t$ represents cell state.
\begin{equation}
	\begin{aligned}
		o_t&=\sigma \left( W_o\left[ h_{t-1},\boldsymbol{\chi }_{t}^{\prime} \right] +b_o \right) \\
		h_t&=o_t\tanh \left( C_t \right) 
		\label{output}
	\end{aligned}
\end{equation}
where $o_t$ is the output gate, responsible for determining which part of memory contributes to the output. $h_t$ is the final output

\emph{\textbf{Spatial Blocks}}. To capture complex spatial dependencies, two distinct spatial blocks are designed: origin-destination spatial block (ODSB), utilizing $$A^{\text{OD}}$$, and semantic spatial block (SESB), employing $A^{\text{SE}}$. This two blocks are formulated as follows:
\begin{equation}
	\begin{aligned}
		H_{t}^{2}&=\text{ODSB}\left( H_{t}^{1},  A^{\text{OD}}\right)\\
		H_{t}^{3}&=\text{SESB}\left( H_{t}^{2}, A^{\text{SE}} \right)
	\end{aligned}
\end{equation}

Both two spatial blocks employ graph attention layers \cite{VPGAAPY2017}, which assign different weights to neighboring nodes. The core of graph attention layers is the attention coefficient $\alpha _{ij}$, calculated as follows:

\begin{equation}
	\begin{aligned}\label{attention}
		\alpha _{ij}&=\frac{\exp \left( \text{Leaky}\mathrm{Re}LU\left( e_{ij} \right) \right)}{\sum_{k\in \mathcal{N} \left( r_i \right)}{\exp \left( \text{Leaky}\mathrm{Re}LU\left( e_{ij} \right) \right)}}\\
		e_{ij}&=a\left( \left[ WH_{t,i}\left\| WH_{t,j} \right. \right] \right) ,r_j\in \mathcal{N} \left( r_i \right)
	\end{aligned}
\end{equation}
where $e_{ij}$ quantifies the importance of grid $r_i$ to its neighbor $r_j$. $\mathcal{N} \left( r_i \right)$ is the set of neighboring grids for grid $r_i$. $a$ denotes the shared attention mechanism that maps high-dimensional features to a scalar value. $W$ is a learnable parameter. 

Subsequently, utilizing $\alpha_{ij}$, the output $H_{t}(K)$ of the graph attention layers is as follows:
\begin{equation}\label{gal_output}
	H_{t}\left( K \right) \,\,=\,\,\underset{k=1}{\overset{K}{||}}\sigma \left( \sum_{r_j\in \mathcal{N} _{r_i}}{\alpha _{ij}^{k}W^kH_{t,j}} \right)
\end{equation}
where $K$ indicates the number of attention heads.

\subsection{Feature decoder module}
Within the feature decoder module, different blocks are designed for distinct subtasks, as shown in Fig \ref{framework} (e). Supervised task adopts a spatial block (SUPB) with graph attention layers, while self-supervised task employs a simpler block (SSB) with fully connected layers, which can be described as follows:

\begin{equation}
	\begin{aligned}
		\hat{Y}_{t}^{\text{sup}}&=\text{SUPB}\left( H_{t}^{3} \right)(\text{supervised task})\\
		\hat{Y}_{t}^{\text{ss}}&=\text{SSB}\left( H_{t}^{3} \right)(\text{self-supervised task})
	\end{aligned}
\end{equation}
where $\hat{Y}_{t}^{\text{sup}}\in \mathbb{R} ^{N\times d_{\text{sup}}}$ and $\hat{Y}{t}^{ss}\in \mathbb{R} ^{N\times d_{\text{ss}}}$ represent the inference results for the supervised task and self-supervised task, with $d_{\text{sup}}$ and $d_{\text{ss}}$ representing their corresponding output dimensions.

{1}

\section{Acknowledgements}
This work is supported by the National Natural Science Foundation of China through grant no. 72071163.
\section{Author contributions}
M.X. and K.H conceived and led the research project. K.H. provides advice on feature and framework design. M.X designed and implemented the framework, producing experimental results. M.X. and K.H. wrote the paper. W.H. and W.J. contributed with ideas to improve the work. All authors reviewed the manuscript.

\appendix
\onecolumn
\setcounter{table}{0}   
\setcounter{figure}{0}
\setcounter{section}{0}
\setcounter{equation}{0}
\renewcommand{\thetable}{Appendix \arabic{table}}
\renewcommand{\thefigure}{Appendix \arabic{figure}}
\renewcommand{\thesection}{Appendix \Alph{section}}
\renewcommand{\thesubsection}{\thesection.\arabic{subsection}}
\renewcommand{\thesubsubsection}{\thesubsection.\arabic{subsubsection}}
\renewcommand{\theequation}{Appendix \arabic{equation}}
\section{Dataset and features} \label{Dataset and features}
\subsection{Air quality data}
\begin{figure*}[h!]
	\centering
	\includegraphics[width=0.8\linewidth]{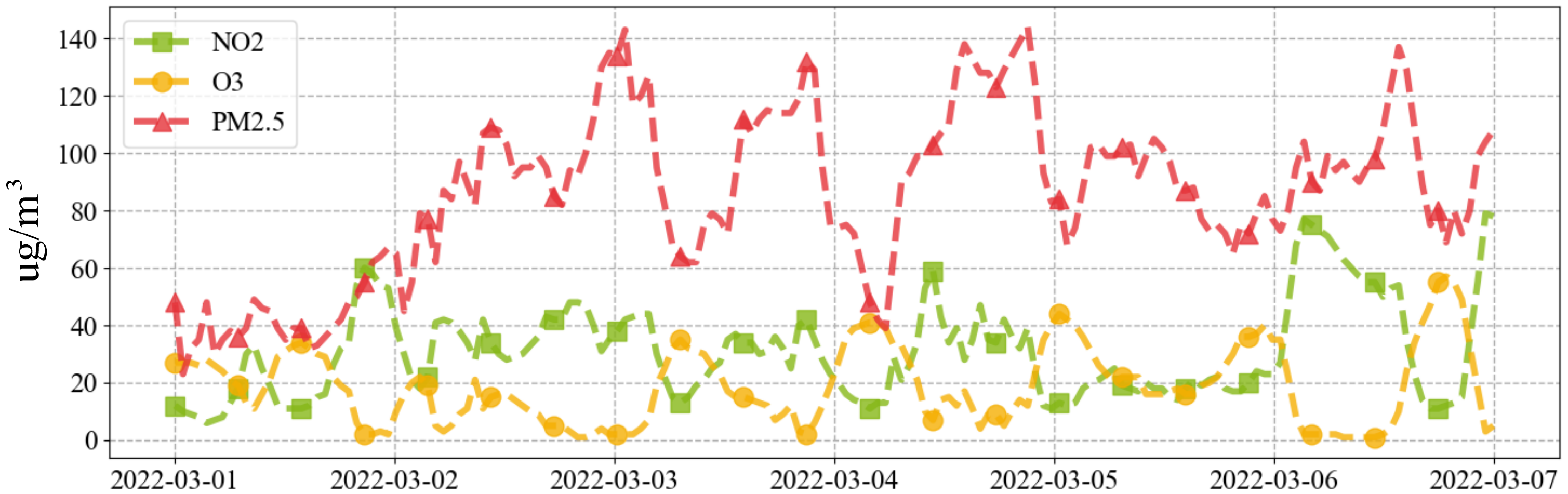}
	\caption{NO$_2$, O$_3$ and PM$_{2.5}$ concentrations during Mar. 1 - 7, 2022}
	\label{NO2_O3_pm2.5}
\end{figure*}

Due to the high-quality of standardized stations, target pollutant concentrations from them are employed as labels for the supervised task. Additionally, chemical reactions between pollutants create correlations in their concentrations, as illustrated in Fig. \ref{NO2_O3_pm2.5}. For example, NO$_2$ and O$_3$ transform into each other under certain meteorological conditions, and NO$_2$ can react chemically with other pollutants to produce PM$_{2.5}$ \cite{HSHYAXFX2011, CLM2001, CP1979}. Consequently, the most relevant pollutant ($Y^{\text{REP}}$) is utilized as feature for air quality inference. However, the spatial sparsity of the standardized stations results in a large number of missing values for raw $Y^{\text{REP}}$. Therefore, IDW is employed to spatially interpolate these values, and the interpolation results are used as the final $Y^{\text{REP}}$.

\subsection{Meteorological data}
\begin{figure*}[h]
	\centering
	\includegraphics[width=1.\linewidth]{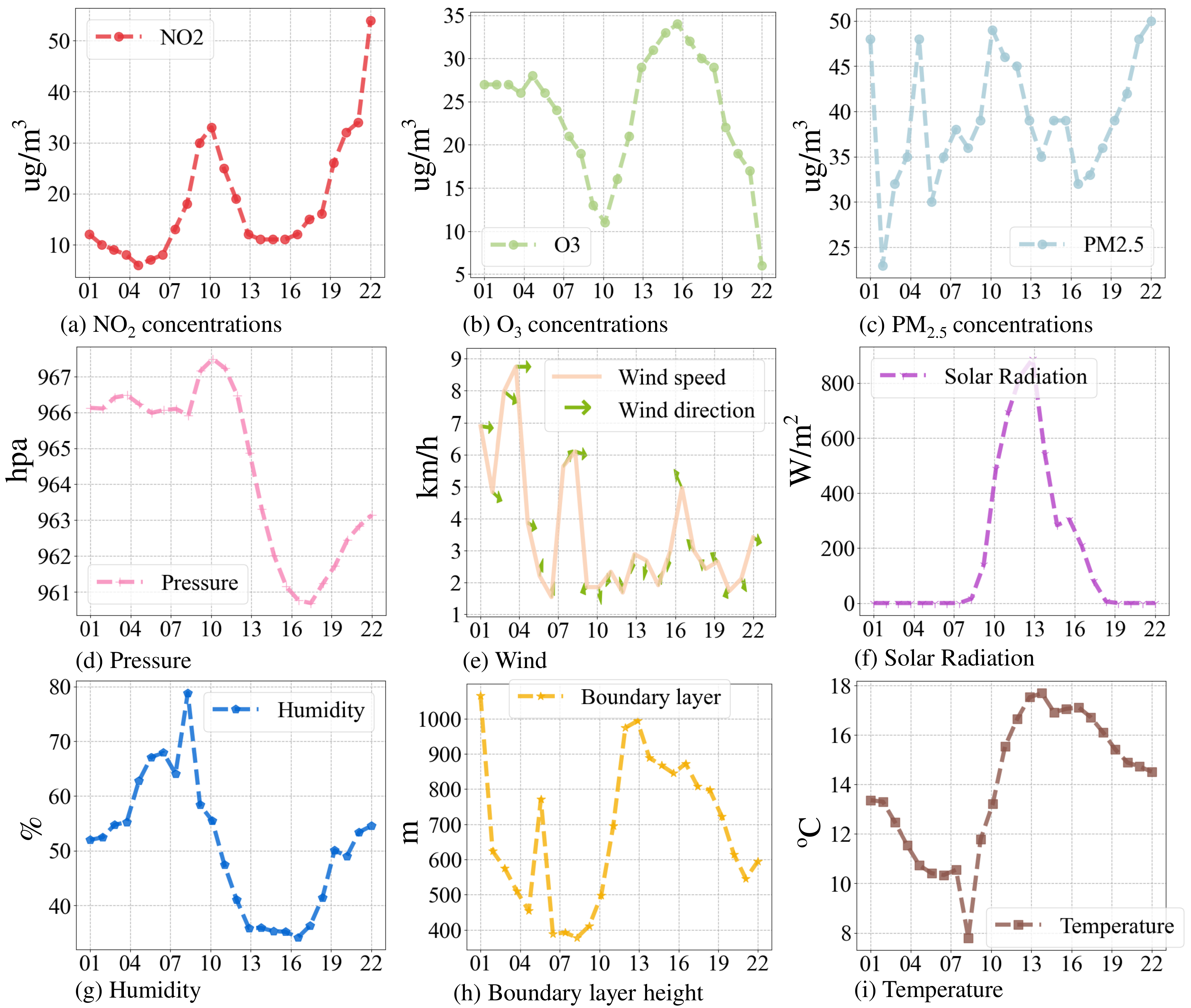}
	\caption{Pollutant concentrations and meteorological conditions during Mar. 1, 2022}
	\label{Meteorological}
\end{figure*}

Previous studies have shown that meteorological conditions significantly affect pollutant concentrations \cite{ZYFH2013, MMMM2023}. The relationship between meteorological conditions and the concentrations of NO$_2$, O$_3$, and PM$_{2.5}$ is further explored in Fig. \ref{Meteorological}. For instance, on Mar. 1, 2023, at 4:00 AM, conditions such as low boundary layer height and solar radiation, medium temperature and humidity, high pressure and wind speed correspond to low NO$_2$ concentrations, high O$_3$ concentrations, and medium PM$_{2.5}$ concentrations. Considering the impact of wind direction on pollutant diffusion direction \cite{SXWHQ2021}, seven meteorological factors ($X^{\text{ME}}$) are finally selected as features for air quality inference: boundary layer height, temperature, humidity, pressure, wind speed, wind direction, and solar radiation.

\subsection{Traffic data} \label{Traffic data}
Traffic has been identified as a major contributor to air pollutant emissions. Specifically, Karagulian \textit{et al.} \cite{KFBCDC2015} show that traffic is responsible for approximately 25\% of $PM$ emissions, and Baldasano \textit{et al.} \cite{BJ2020} note that traffic emissions account for about 65\% of total NO$_2$ emissions. Consequently, three types of traffic-related features are designed for air quality inference: the origin-destination adjacency matrix ($A^{\text{OD}}$), derived from the movement patterns of construction waste trucks; the grid congestion index ($X^{\text{CI}}$), calculated based on road congestion; and construction waste truck flow ($X^{\text{FL}}$), determined from the trajectory points of construction waste trucks.

\begin{figure}[h]
	\centering
	\includegraphics[width=0.3\linewidth]{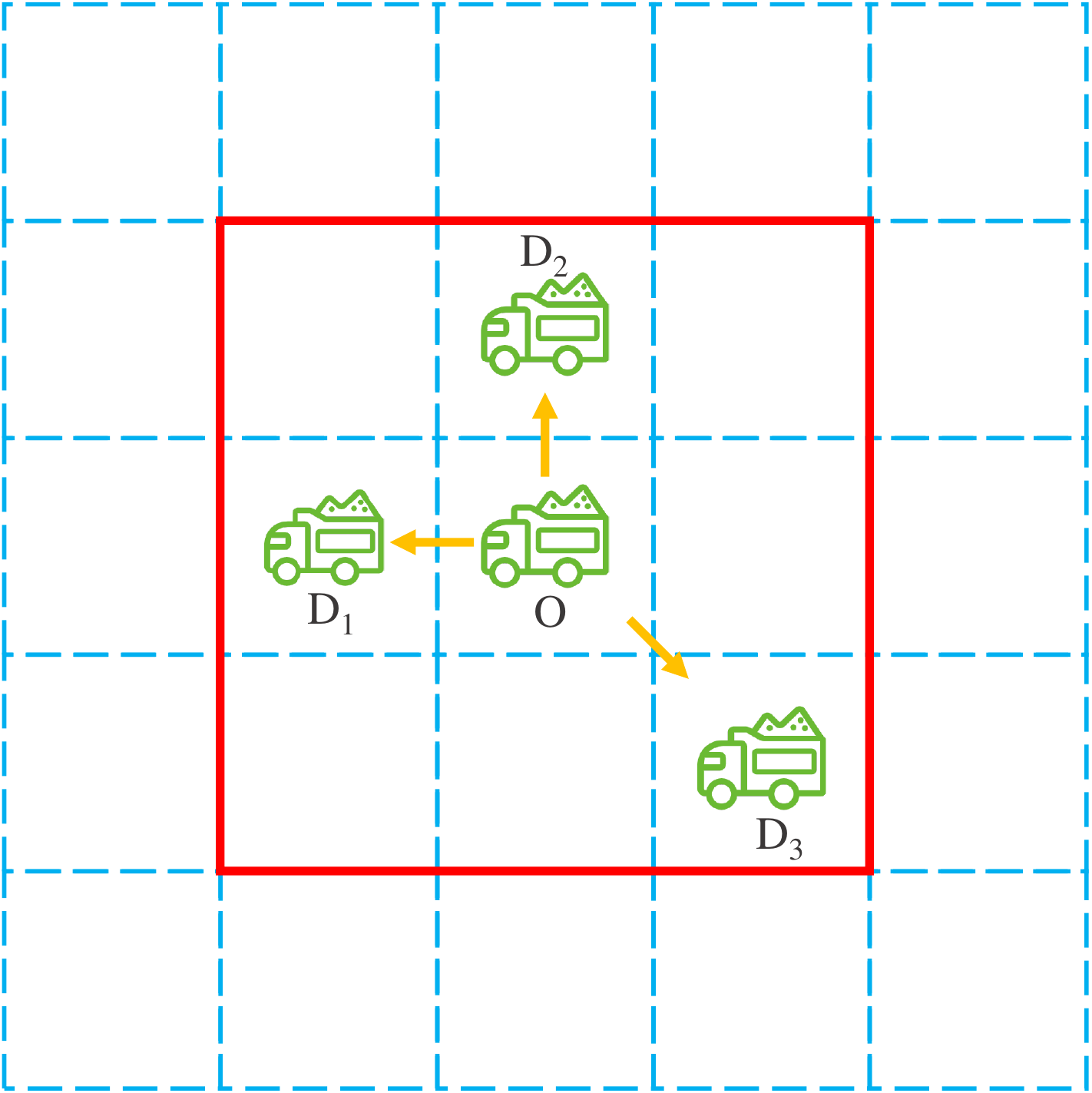}
	\caption{One-hop OD relationship between grids, with the red box indicating regions where this relationship exists.}
	\label{one-hop}
\end{figure}

\textbf{OD adjacency matrix ($A^{\text{OD}}$)}: Due to the lack of modern exhaust treatment systems, heavy-duty diesel vehicles emit more air pollutants than other vehicles \cite{APJJ2011}, making them a major source of traffic air pollutants \cite{KFCCPASHM2015, JPXL2020}. Specifically, in China, heavy-duty diesel vehicles account for 70.4\% of $NO_x$ emissions and 51.9\% of $PM_x$ emissions from all vehicle types \cite{HYCLZLW2023}. Among the heavy-duty diesel vehicles, construction waste trucks are one of the most common vehicles. Consequently, an OD adjacency matrix ($A^{\text{OD}}$) is designed to approximate the spatial dependency of traffic-related pollutant emissions using the movement patterns of construction waste trucks, a type of diesel-powered vehicle. Specifically, the trajectory points of these trucks are employed to obtain the one-hop origin-destination relationship (OHODR) between the grids, as shown in Fig. \ref{one-hop}. Subsequently, $A^{\text{OD}}$ can be defined as follows:

\begin{equation}
	A^{\text{OD}}\left( i,j \right) =\begin{cases}
		1, \text{ if } r_i \text{ and } r_j \text{ have OHODR}\\
		0, \text{ otherwise}\\
	\end{cases}
\end{equation}

\textbf{Construction waste truck flow ($X^{\text{FL}}$)}: Construction waste trucks are known for generating on-road and on-site fugitive dust, which is a significant contributor to atmospheric particulate matter \cite{WHHLTQZFSTHJ2023}.  Consequently, construction waste truck flow ($X^{\text{FL}}$) is used as a feature for PM$_{2.5}$ concentration inference.

\textbf{Grid congestion index ($X^{\text{CI}}$)}: Pollutant emissions escalate on congested roadways due to increased idling and more frequent accelerations of vehicles \cite{AKHAM2002, CCBWMJ2005}. Furthermore, the relationship between pollutant concentration fluctuations and traffic congestion exhibits a positive cross-correlation \cite{SKBKCL2018}. Consequently, the grid congestion index ($X^{\text{CI}}$) is proposed and defined as follows: 

\begin{equation}
	X_{t,i}^{CI}=\frac{\sum_{l_{j,i}\in r_i}{c_{t,j}l_{j,i}}}{\sum_{l_{j,i}\in r_i}{l_{j,i}}}
\end{equation}
where $X_{t,i}^{\text{CI}}$ identifies the grid congestion index of $r_i$ at timestamp $t$, $c_{t,j}$ represents the congestion index of road $j$ at the same time, and $l_{j,i}$ indicates the length of road $j$ within the grid $r_i$.

\subsection{Geographic data} \label{Geographic}
Air pollutant concentrations are closely associated with land usage and function, which can be characterized through geographic data. Consequently, two types of features are designed: geographic features ($X^{\text{GE}}$) and the semantic adjacency matrix ($A^{\text{SE}}$), constructed based on the similarity of $X^{\text{GE}}$.

\textbf{Geographic features $X^{\text{GE}}$}: $X^{\text{GE}}$ contains three main types of features: POI features ($X^{\text{POI}}$), land use features ($X^{\text{LU}}$), and road length features ($X^{\text{RL}}$). More details can be found in Table \ref{Geographic data}.
\begin{table}[h]
	\caption{Details of $X^{\text{POI}}$, $X^{\text{LU}}$ and $X^{\text{RL}}$.}
	\label{Geographic data}
	\begin{tabular}{cc}
		\toprule
		Feature type&Details\\
		\hline
		\multirow{5}{*}{$X^{\text{POI}}$}
		&Culture and education, Financial institutions\\
		&Food and beverage, Shopping malls\\
		&Companies, Transportation spots, Hospitals\\
		&Traveling spots, Hotels, Stadiums\\
		&Residential buildings, Vehicle Services\\
		\cline{2-2}
		\multirow{2}{*}{$X^{\text{LU}}$}
		&Traffic route, Tree cover, Water\\
		&Grassland, Cropland, Buildings, Sparse vegetation\\
		\cline{2-2}
		\multirow{3}{*}{$X^{\text{RL}}$}
		&Trunk, Trunk link, Primary\\
		&Primary link, Secondary\\
		&Secondary link, Tertiary, Tertiary link\\
		\bottomrule
	\end{tabular}
\end{table}

\textbf{Semantic adjacency matrix ($A^{\text{SE}}$)}: Grids with similar geographical features tend to exhibit comparable pollutant concentrations. For example, grids with a similar number of factories might have similar pollutant concentrations. Consequently, semantic adjacency matrix ($A^{\text{SE}}$) is designed. Specifically, K-means \cite{KKM1999} is employed to cluster grids based on their $X^{\text{GE}}$, and $A^{\text{SE}}$ is defined as follows:
\begin{equation}
	A^{\text{SE}}\left( i,j \right) =\begin{cases}
		1, \text{ if } r_i \text{ and } r_j \text{ in the same cluster}\\
		0, \text{ otherwise}\\
	\end{cases}
\end{equation}

\subsection{Timestamp data} \label{Timestamp data}
\begin{figure}[h!]
	\centering
	\includegraphics[width=0.6\linewidth]{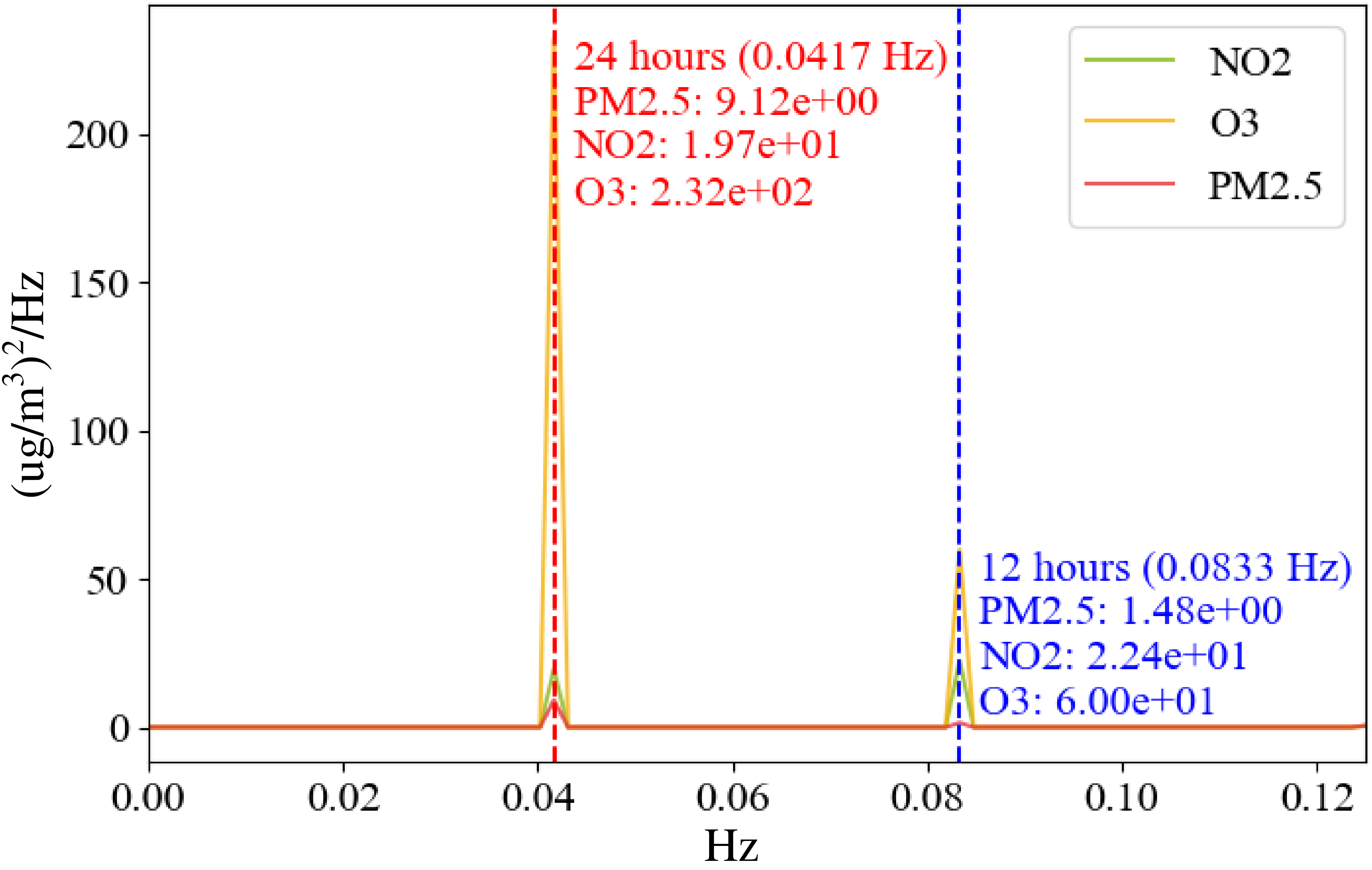}
	\caption{The power spectral density (PSD) of the seasonality for NO$_2$, O$_3$ and PM$_{2.5}$. Each peak denotes a periodicity.}
	\label{PSD}
\end{figure}

Fig. \ref{PSD} presents the power spectral density (PSD) of seasonalities, revealing the partial periodic patterns of pollutant concentrations. Consequently, features representing hour of day ($X^{\text{HD}}$) and day of week ($X^{\text{DW}}$) are extracted from timestamps to capture the daily and weekly patterns respectively, which are commonly employed to infer pollutant concentrations. Notably, $X^{\text{HD}}$ and $X^{\text{DW}}$ are the only two categorical features used in this paper.

\subsection{Data summary}
In conclusion, one type of label and ten types of features are aggregated, which could further divided into 42 subcategories, as detailed in the Table \ref{data}.

\begin{table}[h]
	\caption{Data summary}
	\label{data}
	\begin{tabular}{ccc}
		\toprule
		Type & Name & Element Description \\
		\midrule
		\multirow{8}{*}{$X_{\text{num}}$} &
		$X^{\text{TRL}}$ & Trend data extracted from micro-stations \\
		& \multirow{3}{*}{$X^{\text{REP}}$}& Most relevant pollutant concentrations after spatial interpolation. \\& &For O$_3$, NO$_2$, and PM$_{2.5}$, the most relevant pollutants \\& &are NO$_2$, O$_3$, and NO$_2$, respectively.\\
		& \multirow{3}{*}{$X^{\text{ME}}$}& Boundary layer height $X^{\text{BLH}}$, Temperature $X^{\text{TE}}$ \\
		& &Humidity $X^{\text{HU}}$, Pressure $X^{\text{PR}}$, Wind speed $X^{\textbf{WS}}$\\
		& &Wind direction $X^{\text{WD}}$, Solar radiation $X^{\text{SR}}$\\
		&$X^{\text{CI}}$&Grid congestion index\\
		&$X^{\text{GE}}$& POI, Land use, Road length\\
		&$X^{\text{FL}}$& Construction waste truck flow\\
		\hline
		\multirow{2}{*}{$X_{\text{cat}}$}
		&$X^{\text{HD}}$&Hour of day\\
		&$X^{\text{DW}}$&Day of week\\
		\hline
		\multirow{2}{*}{$\boldsymbol{A}$}
		&$A^{\text{OD}}$ & OD adjacency matrix\\
		&$A^{\text{SE}}$&Semantic adjacency matrix\\
		\hline
		$Y$ & $Y_c$ & Target pollutant concentrations from standardized stations\\
		\bottomrule
	\end{tabular}
\end{table}

\section{Experiments}
\subsection{Experimental Settings} \label{Setting}
\begin{figure*}[h]
	\centering
	\includegraphics[width=1.0\linewidth]{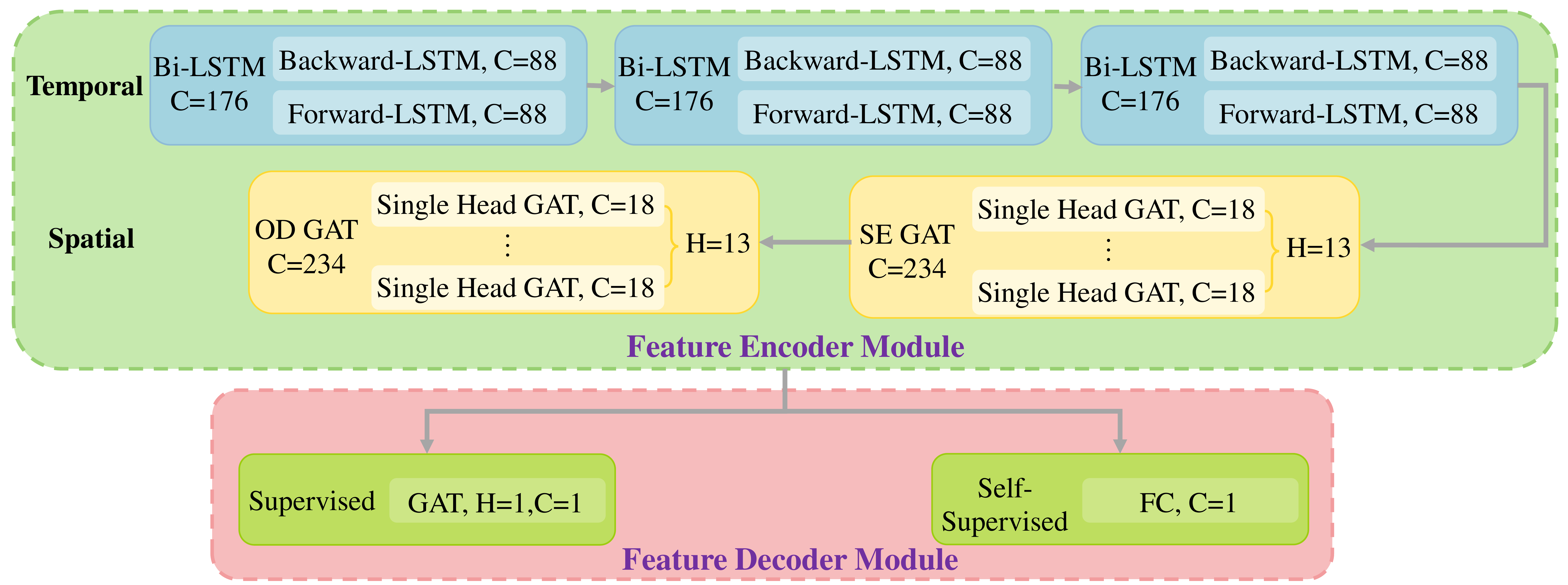}
	\caption{Hyperparameter settings of MTSTN.}
	\label{parameter-set}
\end{figure*}

Fig. \ref{parameter-set} report the detail hyperparameter settings of MTSTN's architecture.

\subsection{Hyperparameter Study} \label{Hyperparameter study}
A systematic investigation into the performance of MTSTN under different hyperparameter settings is conducted. In each experiment, a single hyperparameter is modified while the others remain unchanged as the default settings, thereby isolating the effects of other parameters. Notably, although each hyperparameter is incrementally increased from its minimum value, only the critical results are presented in this paper.

\begin{figure*}[h]
	\centering
	\includegraphics[width=1.0\linewidth]{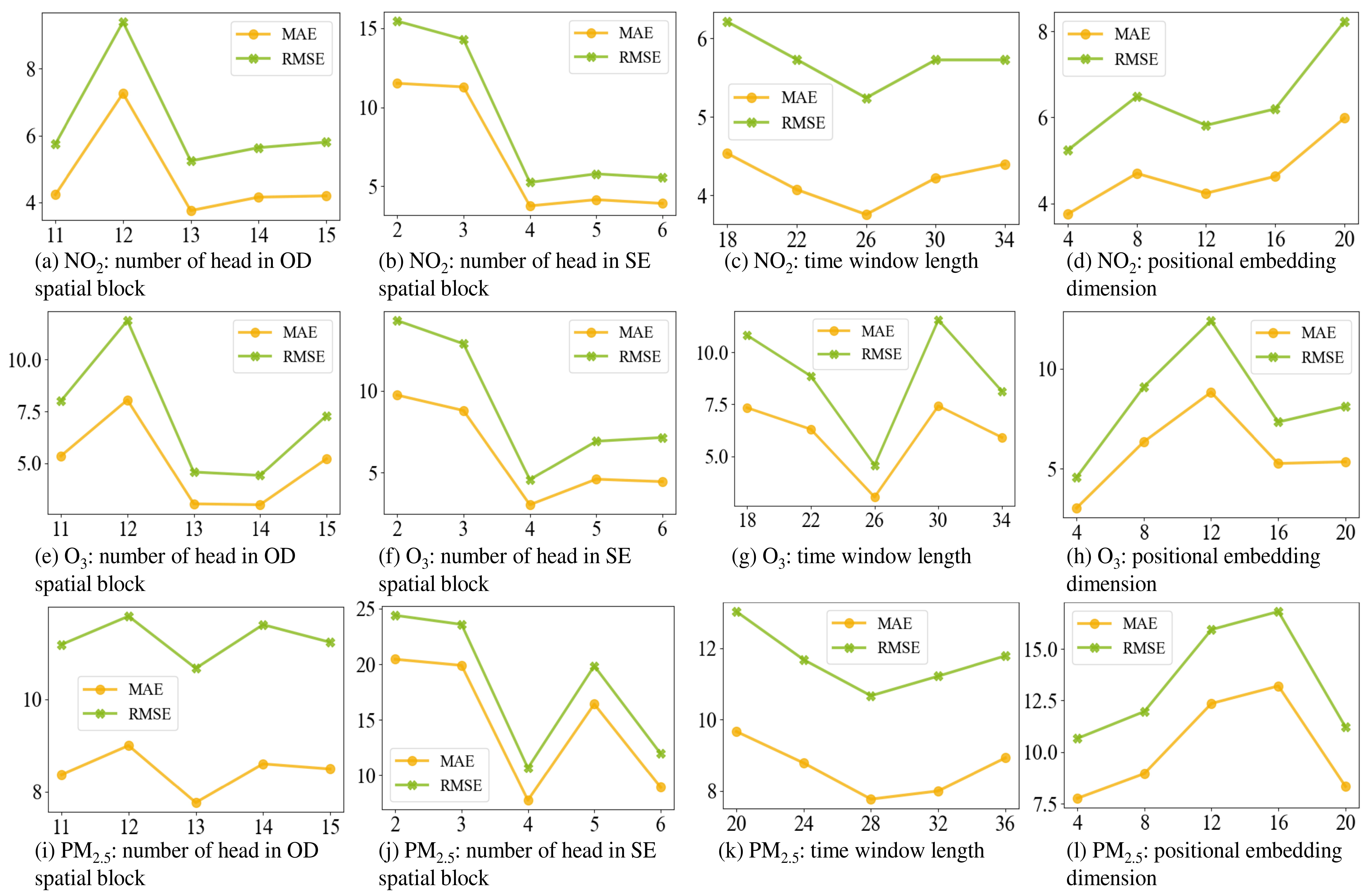}
	\caption{The performance of MTSTN under different parameter settings.}
	\label{Hyperparameter}
\end{figure*}

\subsubsection{Effects of number of head in graph attention layers} The number of heads in the graph attention layers of the OD and SE spatial blocks are adjusted, and the results for MAE and RMSE are reported in Fig. \ref{Hyperparameter} (a)-(b), (e)-(f), and (i)-(j). When MTSTN's performance peaks, OD spatial block employs more heads than SE spatial block, suggesting greater complexity within $A^{OD}$.

\subsubsection{Effects of time window length} As shown in Fig. \ref{Hyperparameter} (c), (g), and (k), the errors of MTSTN exhibit a downward trend until $\tau$ reaches 26 or 28, after which they begin to increase. This phenomenon suggests that longer time windows cannot guarantee better results.

\subsubsection{Effects of positional embedding dimension} Finally, the positional embedding dimension $d_p$ is modified. As shown in Fig. \ref{Hyperparameter} (d), (h), and (l), an increase in $d_p$ leads to a decline in MTSTN's performance. This decline may be attributed to the increased distance between data points in higher dimensions, which hinders MTSTN’s ability to capture the data's local structure.

\subsection{Baselines}
MTSTN is compared with nine baselines, which can be categorized into interpolation models, machine learning models, and deep learning models. Further details about the baselines are provided below.
\begin{itemize}
	\item {\textbf{KNN}}: K-Nearest Neighbors estimates the unknown nodes by interpolating the readings of the K closest nodes \cite{HJYZLYR2023}.
	
	\item {\textbf{IDW}}: Inverse Distance Weighting is a deterministic interpolation method that infers unknown nodes using the weighted averages of nearby available nodes, with closer nodes having more influence \cite{LGD2008}.
	
	\item {\textbf{LUR}}: Land Use Regression utilizes related exogenous covariates, such as land use and traffic characteristics, to estimate pollutant concentrations for unknown nodes \cite{HDOCCML2014}.
	
	\item {\textbf{RF}}: Random forest is extensively used and recognized for its efficient performance in handling non-linear regression tasks \cite{FKME2014}.
	
	\item {\textbf{XGBoost}}: Extreme Gradient Boosting is an efficient and scalable machine learning algorithm known for its superior performance in both classification and regression tasks \cite{CTGC2016}.
	
	\item {\textbf{STGCN}}: Spatio-Temporal Graph Convolutional Networks is a purely convolutional structure that enables faster training with fewer parameters \cite{YBHZ2017}.
	
	\item {\textbf{MSTGCN}}: Multi-View Spatial-Temporal Graph Convolutional Networks not only capture the most relevant spatio-temporal information but also exhibit strong domain generalization capabilities \cite{JZYJXYRYH2021}.
	
	\item {\textbf{ASTGNN}}: Attention-based Spatial-Temporal Graph Neural Network efficiently captures temporal dynamics, spatial dynamics, and spatial heterogeneity through a novel self-attention mechanism \cite{GSYHX2021}.
	
	\item {\textbf{PDFormer}}: Propagation Delay-Aware Dynamic Long-Range Transformer is one of the state-of-the-art spatio-temporal prediction model, capable of capturing delayed propagation \cite{JJCWJ2023}.
	
\end{itemize}

\end{document}